\documentclass[conference]{IEEEtran}
\IEEEoverridecommandlockouts
\usepackage{cite}
\usepackage{amsmath,amssymb,amsfonts}
\usepackage{algorithmic}
\usepackage{graphicx}
\usepackage{textcomp}
\usepackage{xcolor}
\usepackage{color}
\usepackage{caption}
\usepackage{subcaption}

\def\BibTeX{{\rm B\kern-.05em{\sc i\kern-.025em b}\kern-.08em
    T\kern-.1667em\lower.7ex\hbox{E}\kern-.125emX}}
\begin{document}

\title{Object 3D Reconstruction based on Photometric Stereo and Inverted Rendering}

\author{\IEEEauthorblockN{{1\textsuperscript{st} Anish R. Khadka}
\textit{Kingston University London}\\
London, UK \\
A.Khadka@kingston.ac.uk}
\and
\IEEEauthorblockN{{2\textsuperscript{nd} Paolo Remagnino}
\textit{Kingston University London}\\
London, UK \\
P.Remagnino@kingston.ac.uk}
\and
\IEEEauthorblockN{{3\textsuperscript{rd} Vasileios Argyriou}
\textit{Kingston University London}\\
London, UK \\
Vasileios.Argyriou@kingston.ac.uk}
}

\maketitle

\begin{abstract}
Methods for 3D reconstruction such as Photometric stereo recover the shape and reflectance properties using multiple images of an object taken with variable lighting conditions from a fixed viewpoint. Photometric stereo assumes that a scene is illuminated only directly by the illumination source. As result, indirect illumination effects due to inter-reflections introduce strong biases in the recovered shape. Our suggested approach is to recover scene properties in the presence of indirect illumination. To this end, we proposed an iterative PS method combined with a reverted Monte-Carlo ray tracing algorithm to overcome the inter-reflection effects aiming to separate the direct and indirect lighting. This approach iteratively reconstructs a surface considering both the environment around the object and its concavities. We demonstrate and evaluate our approach using three datasets and the overall results illustrate improvement over the classic PS approaches.
\end{abstract}

\begin{IEEEkeywords}
Photometric Stereo, 3D Reconstruction, Ray Tracing
\end{IEEEkeywords}

\section{Introduction}
\label{sec:intro}

Scene and object 3D reconstruction is the process of capturing their shape and appearance using various methods and approaches such as stereo, structure from motion, shape from shading, and many more \cite{Remondino2006Imagebased3M}. The reconstruction is highly applicable in a number of fields as it provides the ability to understand 3D scenes and objects on basis of 2D  images. The applications ranging from robotics and automated industrial quality inspection over human-machine interaction \cite{6977392} (example action, gesture and face recognition), satellite 3D data analysis \cite{7563843}, to movies and architectural applications \cite{Herbort2011AnIT}. Additionally, the method is commonly used to analyse the surfaces of a celestial object, such as the Moon \cite{Hicks2011APF}.

Photometric stereo (PS) is a well-established technique that is used for 3D surface reconstruction \cite{Esteban2008MultiviewPS}. The approach generally inherits the principle of appearance analysis of a 3D object on its 2D images. Based on the intensity information, these approaches attempt to infer the shape of the depicted object \cite{Herbort2011AnIT}. It estimates shape and recovers surface normals of a scene by utilising several intensity images obtained under varying lighting conditions with an identical viewpoint  \cite{Tankus2005PhotometricSU,Hayakawa2002PhotometricSU}. By default, PS assumes a Lambertian surface  reflectance; a standard reflectance model which defines a linear dependency between the normal vectors and image intensities. The definition of the model then can be used to determine the 3D space in the image \cite{Belhumeur1996WhatIT}.  However, just a single Lambertian image is not adequate to correctly determine the surface shape. Therefore, the PS uses several images whose pixels corresponds to a single point on the object and is able to recover surface normals and albedos \cite{Tan2008SubpixelPS}.

Light displays complicated attributes while interacting with objects resulting direct and indirect illumination as shown in figure \ref{nbounceImage}.However, classical PS naively assumes that a scene is illuminated only directly by the emitting source. In presence of indirect illumination, it produces erroneous results with reduced reconstruction accuracy \cite{Ikeuchi1981DeterminingSO}. For example, an indirect illumination such as inter-reflections makes concave objects appear shallower \cite{Nayar1990ShapeFI}.

In this paper, we present an iterative 3D reconstruction method considering inter-reflections due to the concavities and the environment. We propose a novel method that accounts for inter-reflections in a calibrated photometric stereo environment. This approach utilises a reverted Monte Carlo ray tracing method to extract the environmental colour trying to minimise the inter-reflections within images used for photometric stereo. This approach not only accommodates the concave surface but also applies to any object in a scene with inter-reflections. The proposed method Iterative Ray Tracing Photometric Stereo - IRT PS iteratively applies Photometric Stereo (PS) and a reverted ray tracing algorithm based on a Monte-Carlo implementation to reconstruct with higher accuracy the observed surfaces. This approach iteratively reconstructs the surface and separates the indirect from direct lighting considering also the environment around the object.  Likewise, the proposed IPT-PS method can be integrated to any PS technique removing the effects of inter-reflections and improving the overall reconstruction accuracy.

Our approach is extensively evaluated on three datasets and the overall results demonstrate improvement over the classic approaches. The main contributions of our work are:

\begin{itemize}
 \item a reverted Monte Carlo ray tracing algorithm to estimate the indirect lighting both from the environment and the object's concavities.
 \item an iterative surface reconstruction method that is utilised by the reverted Monte Carlo ray tracing

 \item the proposed methodology that allows IRT-PS to be combined with any other PS algorithm improving the overall performance.
\end{itemize}

The paper is organised as follows: Section \ref{photometricStereoLable} provides background material on Photometric Stereo, followed by invert light transport, their properties and related works. In section 3, we introduce the mathematical definition of necessary the terms. In section 4, we propose a novel iterative PS method and discuss the suggested reverted Monte Carlo ray tracing algorithm. The performance of this approach is investigated in section 5, with section 6 concluding the work.


\section{Photometric Stereo}\label{photometricStereoLable}

Photometric stereo (PS) is an approach to estimate the surface normal and reflectance (i.e albedo) of an object based on three or more intensity images with the fixed view under varying lighting condition \cite{Hayakawa2002PhotometricSU}. A number of solutions have been proposed to address this problem. Woodham \cite{Woodham1978} was the first to introduce the PS method. He proposed an approach simple and effective. However, he only considered Lambertian surface which suffers from noise. In his method,  it is assumed that the surface albedo is known a prior for each point on the surface, the surface gradient can be obtained by using a three-point light source. Onn and Bruckstein \cite{Onn1990IntegrabilityDS} developed a two-image PS method. Their work was based on the assumption that the objects are smooth and no self-shadows are present. The PS was further extended by Coleman and Jain \cite{Coleman1982}, which utilises four light sources, discards the specular reflections and estimates the surface shape by performing mean of diffuse reflections and the use of the Lambertian reflection model. Nayar \textit{et al.} \cite{Nayar1990} proposed a PS method which used a linear combination of an impulse specular component and the Lambertian model to recover the shape and reflectance for a surface. Similarly, an algorithm for estimating the local surface gradient and real albedo from four sources in the presence of highlights and shadows was proposed by  Barsky and Petrou \cite{Barsky2001b}. Chandraker \textit{et al.}\cite{Chandraker2007} proposed an algorithm that required at least four light sources and images to reconstruct surface in presence of shadow.  It is also worth mentioning the related work presented in \cite{Levine2005,Finlayson2004,ArgyriouChapter,Ragheb2003} following similar architectures and approaches.

Furthermore, over the previous years, methods that consider images produced by more general lighting conditions not known a prior. Basri \textit{et al.}  \cite{BasriJacobsKemelmacher_IJCV07} proposed a PS method where no prior knowledge of the light source and its type is required, however the emitting source should be distant or unconstrained. They utilised low order spherical harmonics and optimised it to low-dimensional space to represent Lambertian objects. Likewise, Shi \textit{et  al.} \cite{Shi2010SelfcalibratingPS} used colour and intensity profiles, which are obtained from registered pixels across images to propose a self-calibrating PS method. They automatically determine a radiometric response function and resolved the generalised bas-relief for estimating surface normals and albedos. While lighting conditions could be unknown, they required fixed viewpoint.

Nevertheless, a majority of the methods and models while
working well with with matte objects, under-perform when the reconstructed objects are specular, transparent or with inter-reflections. Non-Lambertian reflection and specifically inter-reflection may be difficult to solve in photometric stereo. Solomon and Ikeuchi \cite{Solomon1996} developed a method where they utilised four lights and tried to extract the surface shape and roughness of an object which has specular lobe. They used a simplified version of Torrance-Sparrow reflectance model to determine the surface roughness. Bajcsy \textit{et al.} \cite{Bajcsy1996DetectionOD}  presented an algorithm for detecting diffuse and specular interface reflections and some inter-reflections. They used brightness, hue, and saturation values instead of RGB as they point out that the values have a direct correspondence to body colours and to diffuse and specular, shading, shadows and inter-reflections. But, the algorithm requires uniformly coloured dielectric surface under single coloured scene illumination.
Tozz \textit{et al.} \cite{Tozza2016DirectDP} proposed a PS method that is independent of the albedo values and uses image ratio formulation. However, their method requires an initial separation of diffuse and specular components.

In addition, because of the nature of light, inter-reflection is unavoidable even in a controlled environment. This may vary in magnitude depending on the environment itself, the structure, and the material of the object. Moreover, it may not be uniform over the whole surface. As a result, the images are blurred locally in shade.  Most photometric methods do not consider inter-reflection from an environment and concave surfaces, and those that do have considered one of the two cues only. One of the first attempts at scene recovery under inter-reflection was purpose by Nayar \textit{et al.} \cite{Nayar1990ShapeFI}. They presented an iterative algorithm which recovers shapes from a concave surface which first estimates the shape from intensity data; then this shape is used as input, and the radiosity method is applied to estimate a corrected, no-interreflection image intensity distribution. These steps are carried iteratively until convergence. Nevertheless, the algorithm only examines the inter-reflection in concave shapes, Lambertian reflectance models and does not take into account the colour of the inter-reflected light. Funt and Drew \cite{Funt1993ColorSA} proposed an algorithm which is based on singular value decomposition of the colour for a convex surface. They proposed a \textit{``one-bounce''} model which measured inter-reflection between two matte convex surfaces with a uniform colour and illumination can vary spatially in its intensity but not in its spectral composition. Again, the algorithm is specific to convex surface assuming a uniform colour and illumination that can vary spatially. Langer \cite{Langer1999WhenSB} did the study on the shadows which becomes inter-reflections. They proposed a method for inferring surface colour in a uni-chromatic scene which is based on the relative contrast of the scene in different colour channels. Again, the method is highly specific and only deals with inter-reflection related to shadow.

Most existing shape from intensity techniques accounts for an only direct component of light transport. Nayar \textit{et al.} \cite{Nayar2006FastSO} proposed using high-frequency illumination patterns to separate direct and indirect illumination from more general scenes.Gupta \textit{et al.} \cite{Gupta2009DeFO} studied the relation between illumination defocus and global light transport.  Again, Chen \textit{et al.} \cite{Chen2008ModulatedPF} used modulated structured light patterns with high-frequency patterns to mitigate the effects of indirect illumination. Lamond \textit{et al.} \cite{Lamond2009DiffuseSpeuclar} used high-frequency light patterns to separate the diffuse and specular components of BRDF. Holroyd \textit{et al.} \cite{Holroyd2010ACO} constructed a high-accuracy imaging system for measuring the surface shape and BRDF. All these techniques either are an active method or they assume that the indirect illumination in each of the acquired images is caused by a single source. In contrast, we consider separation of indirect components by simulating the inter-reflections and removing it from the source images.

\section{Forward Light Propagation}

An image captured by the camera is the results of a complex sequence of reflections and inter-reflections. When light is emitted from the source, it bounces off the scene's surface one or more times before reaching to a camera.

\begin{figure}[ht]
\begin{center}
  \includegraphics[width=.95\linewidth]{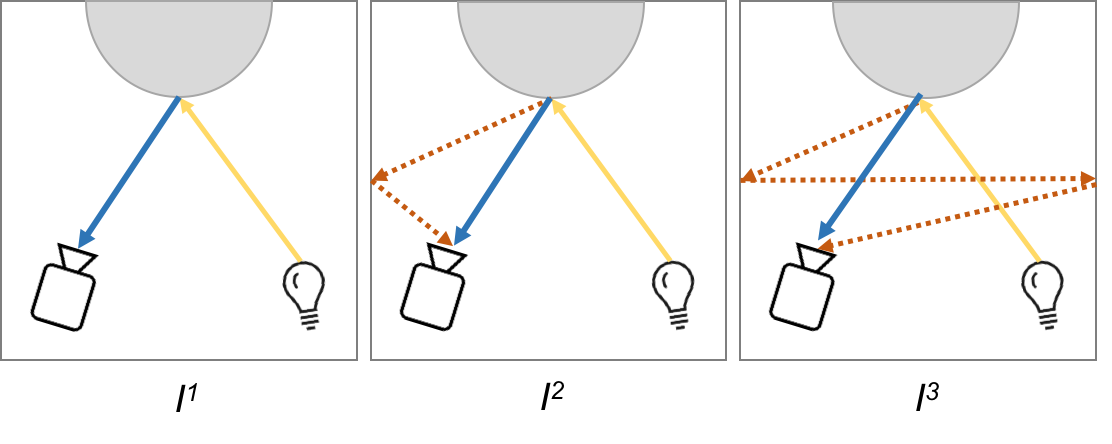}
\end{center}
\caption{ (Left)Direct and (Middle)(Right)indirect light bounce around the environment}
\label{nbounceImage}
\end{figure}

In theory, every image can be captured as infinity sum, $I = I^1 + I^2 + I^3 + .. + I^n$, where $I^n$ denotes the total contribution of light that bounces  $n$ times before reaching the camera as shown in figure \ref{nbounceImage}. For example, $I^1$ is the captured image if it was possible to remove all the indirect illumination from reaching the camera sensor,  while the infinite sum $I^2 + I^3 + .. + I^n$ describes the total contribution of indirect illumination. Although we can capture the final image $I$ using a camera, the individual \textit{“n-bounce”} images are not directly measurable in the real-world scenario.

Nevertheless, the techniques for simulating inter-reflections and other light transport effects are not new in the computer vision and graphics.
The algorithm that simulated the forward light transport was solved by Kajiya \cite{Kajiya1986TheRE}. The algorithm is also known as \textit{rendering equation}. The rendering equation is an integral in which the radiance leaving a point is given as the sum of emitted plus reflected radiance under a geometric optics approximation.

\begin{equation}
\label{renderingEquation} I(x,x')=g(x,x')\Bigg [e(x,x') + \int\limits_s p(x,x',x'')I(x',x'')dx'' \Bigg]
\end{equation}

 Where $I(x,x')$ is related to the intensity of light passing from $x'$ to point $x$. $g(x,x')$ is a "geometry" term, $e(x,x')$ is related to the intensity of emitted light from $x'$ to $x$ and $p(x,x'x'')$ is related to the intensity of light scattered from $x''$ to $x$ by a patch of surface at $x'$.

An algorithm such as ray tracing \cite{Foley1990ComputerG}\cite{Jarosz2008AdvancedGI} solved the equation  \ref{renderingEquation} by using Monte-Carlo methods, whereas radiosity \cite{Foley1990ComputerG}\cite{Immel1986ARM} used finite element method to produce near realistic looking images in the field.

For a Lambertian object illuminated by a light source of parallel rays, the observed image intensity $\mathbf{a}$ at each pixel is given by the product of the albedo $\rho$ and the cosine of the incidence angle $\theta_{i}$ (the angle between the direction of the incident light and the surface normal) \cite{Horn1977}. The above incidence angle can be expressed as the dot product of two unit vectors, the light direction $\mathbf{l}$ and the surface normal $\mathbf{n}$, $\mathbf{a}=\rho \cos(\theta_{i})=\rho (\mathbf{l}\cdot \mathbf{n})$.

Let us now consider a Lambertian surface patch with albedo $\rho$ and normal $\mathbf{n}$, illuminated in turn by several fixed and known illumination sources with directions $\mathbf{l}^{1}$, $\mathbf{l}^{2}$, ..., $\mathbf{l}^{\tilde{Q}}$. In this case we can express the intensities of the obtained pixels as:
\begin{equation}
\label{Eq:R01_PS_Q} \mathbf{a}^{k}=\rho(\mathbf{l}^{k}\cdot \mathbf{n}),\ \ \ \rm{where} \ \  k=1,2,...,\tilde{Q}.
\end{equation}
We stack the pixel intensities to obtain the pixel intensity vector \\ $\mathbf{A}_{a}=(\mathbf{a}_{1},\mathbf{a}_{2},...,\mathbf{a}_{\tilde{Q}})^{T}$. Also the illumination vectors are stacked row-wise to form the illumination matrix $\mathbf{L}=(\mathbf{l}^{1}, \mathbf{l}^{2},...,\mathbf{l}^{\tilde{Q}})^{T}$. Equation~(\ref{Eq:R01_PS_Q}) could then be rewritten in matrix form:
\begin{equation}
\label{Eq:R01_PS_QM} \mathbf{A}_{a}=\rho \mathbf{L} \mathbf{n}
\end{equation}
If there are at least three illumination vectors which are not coplanar, we can calculate $\rho$ and $\mathbf{n}$ using the Least Squares Error technique, which consists of using the transpose of $\mathbf{L}$, given that $\mathbf{L}$ is not a square matrix:
\begin{equation}
\label{Eq:R01_PS_QMInv} \mathbf{L}^{T}\mathbf{A}_{a}=\rho \mathbf{L}^{T}\mathbf{L} \mathbf{n} \Rightarrow (\mathbf{L}^{T}\mathbf{L})^{-1}\mathbf{L}^{T}\mathbf{A}_{a}=\rho \mathbf{n}
\end{equation}
Since $\mathbf{n}$ has unit length, we can estimate both the surface normal (as the direction of the obtained vector) and the albedo (as its length). Extra images allow one to recover the surface parameters more robustly.

\section{Proposed Iterative Ray Tracing Photometric Stereo Method (IRT-PS)}

\begin{figure}[ht]
\begin{center}
  \includegraphics[width=0.3\linewidth]{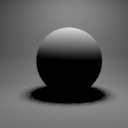}
  \includegraphics[width=0.3\linewidth]{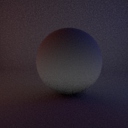}
  \includegraphics[width=0.3\linewidth]{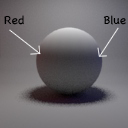}\\
  \subcaption{(Left)Image with no inter-reflection, (Middle) Image with inter-reflection from Environment only, (Right) Combined Image}
   \includegraphics[width=0.3\linewidth]{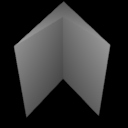}
  \includegraphics[width=0.3\linewidth]{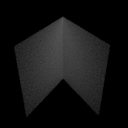}
  \includegraphics[width=0.3\linewidth]{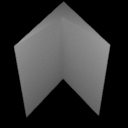}\\
  \subcaption{(Left)Image with no inter-reflection, (Middle) Image with inter-reflection from Concavity only, (Right) Combined Image}
  \end{center}
  \caption{Example images of Inter-reflection from environment and concavity}
  \label{innerreflectionExample_1}
\end{figure}

In nature, when we illuminate a surface, light not only reflects towards the viewer but also among all surfaces in the environment. This is always true, with exception of scenes that consists only of a single convex surface. In general, scenes include concave surfaces where points reflect light between themselves. Furthermore, inter-reflections can occur due to the environment and appreciably can alter a scene's appearance. In figure \ref{innerreflectionExample_1}, to simulate the inter-reflections the sphere is placed within the Cornell box \cite{Niedenthal2002} and highlights the inter-reflections i.e sphere receive the colours from its environment.

Existing computer vision algorithms do not account for effects of inter-reflections and hence often produce erroneous results. The algorithms that are directly affected by inter-reflections are the shape-from-intensity algorithms including Photometric Stereo. Due to the common assumption of single surface reflections (direct illumination) and disregarding higher order (inter-reflections, a subset of global illumination), photometric methods produce erroneous results when applied to open scenes.

\begin{figure}[ht]
\begin{center}
\includegraphics[width=0.95\linewidth]{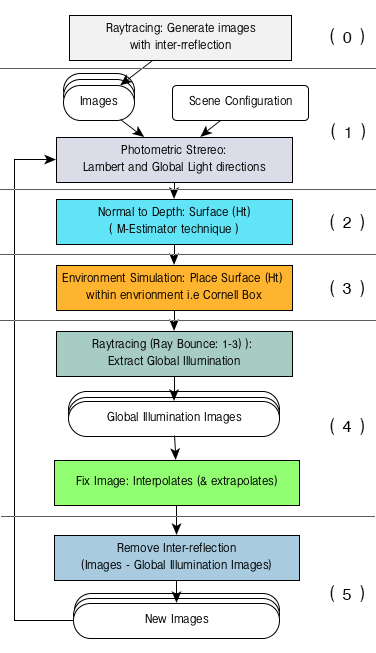}
\end{center}
  \caption{An overview of the proposed IRT-PS algorithm.}
  \label{OverviewOfAlgorithm}
\end{figure}

The first stage of this approach (stage 0), is performed only once throughout the process and involves the acquisition of the initial input images. It is assumed that inter-reflections are present and that the captured surface is within the known environment. In our case within a Cornell Box.

Moving to the following stage, PS is applied to the images acquired at stage 0 using equation \ref{Eq:R01_PS_QMInv} to obtain the initial albedo $\rho_{t}$ and normals $\mathbf{n}_{t}$. Integrating over the obtained normals a 3D surface $H_{t}$ is obtained using the M-estimator technique. This initial surface that is affected by the presence of the inter-reflections becomes the input to the following stage, that involves the proposed reverted ray tracing algorithm.

As environment information is known prior to reconstruction, we can implement our environment. The Cornell Box was setup as the environment at the following stage 3. More realistic textures can be used for the walls without affecting the proposed methodology.

\begin{figure}[ht]
\begin{center}
  \includegraphics[width=0.35\linewidth]{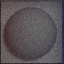}
  \includegraphics[width=0.35\linewidth]{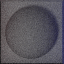}\\
  \includegraphics[width=0.35\linewidth]{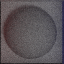}
  \includegraphics[width=0.35\linewidth]{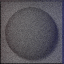}\\
  \end{center}
\caption{Sample images from Stage 0 with inter-reflections due to the environment}
\label{chart:overallErrorChart}
\end{figure}

In stage 4, we simulate the environment assuming the Cornell box is given or estimated. In our case, this approach can be extended to other realistic environmental projection such as  Hemispherical Dome Projection \cite{Swinburne2005SphericalM} without affecting the proposed methodology. Then we place the generated $H_{t}$ surface within this environment.

In the following stage, based on equation \ref{eq:renderingEquationWithMonteCarlo}, the reverted ray tracing algorithm is applied. Since we are only interested in inter-reflections, only the indirect illumination is calculated.To implement the ray tracer for Lambertian surface, we solve the rendering equation by integrating Monte Carlo estimator

\begin{equation}
    L_{0}(p,w_{o}) = \int\limits_\Omega f(p,w_{0},w_{i})L_{i}(p,w_{i})cos\theta_{i}dw_{i}
      \label{eq:renderingEquation}
\end{equation}

Where $L_{0}(p,w_0)$ is the total outgoing radiance reflected at $p$ along the $w_0$ direction. $L_i(p,w_i)$ is the radiance incident at $p$ along the $w_i$ direction. $f(p,w_0,w_i)$ determines how much radiance is reflected at $p$ in direction $w_0$, due to irradiance incident at $p$ along the $w_i$ direction.  $cos\theta_i$ is from the Lambert's cosine law: diffuse reflection is directly proportional to $cos(\theta)$ of the normals and the incident illumination ($i$). Finally, $\int\limits_\Omega  dw_{i}$ is an integral over a given hemisphere.

As Monte-Carlo approximation is a method to approximate the expectation of a random variable, using samples.
\begin{equation}
E(X) \approx \frac{1}{N}\sum^{n}_{i=1} X_i
\end{equation}

where, $E(X)$ is an approximation of average value of random variable $X$.$N$ is the sample size. And when we integrate it to equation \ref{eq:renderingEquation} we solve the rendering equation.

\begin{equation}
    \langle L_{0}(p,w_{o})  \rangle= \frac{1}{N} \sum^{N}_{i=1}\frac{f(p,w_{0},w_{i})L_{i}(p,w_{i})cos\theta_{i}dw_{i}}{p(w_i)}
    \label{eq:renderingEquationWithMonteCarlo}
\end{equation}

However, Monte-Carlo estimator is affected by noise, the ray tracer algorithm also inherited such a problem.  For example, to half the noise in an image rendered by ray tracing, we need to quadruple the number of samples.

\begin{figure}[ht]
\begin{center}
  \includegraphics[width=0.3\linewidth]{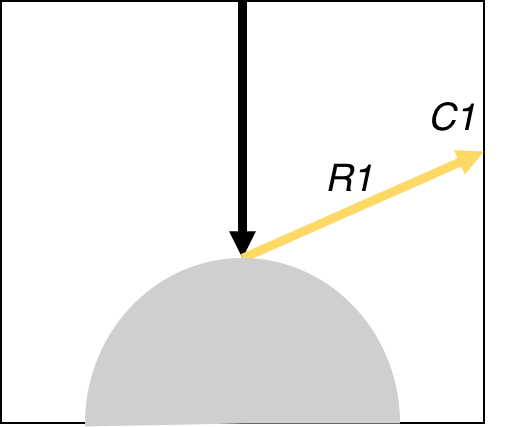}
  \includegraphics[width=0.3\linewidth]{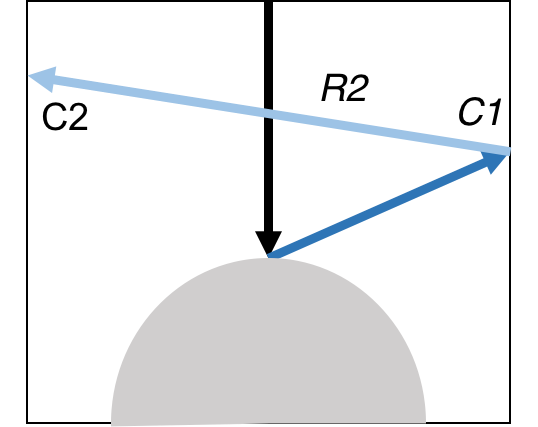}
  \includegraphics[width=0.3\linewidth]{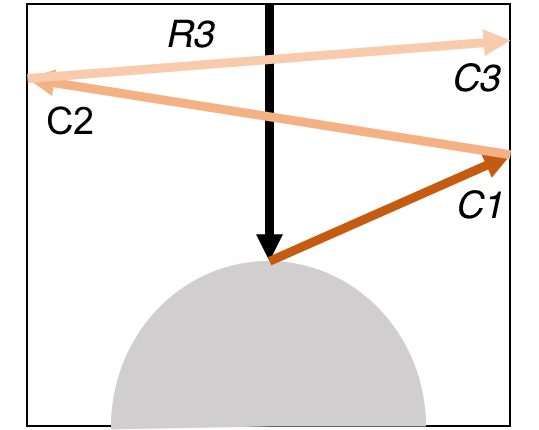}
  \end{center}
\caption{Extraction of Environment Intensities in 3 different ways (a) Only extract colour (c1), (b) reflect ray one time and combine the intensities (c1 * c2), and (c), reflect one more time and combine all the colours (c3*c2*c1).}
\label{fig:environmentColourExtraction}
\end{figure}

\begin{figure}[ht]
\begin{center}
  \includegraphics[width=0.30\linewidth]{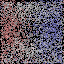}
  \includegraphics[width=0.30\linewidth]{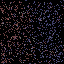}
  \includegraphics[width=0.30\linewidth]{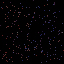}\\
  \subcaption{Environment Colour extracted for Sphere}
  \includegraphics[width=0.30\linewidth]{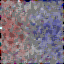}
  \includegraphics[width=0.30\linewidth]{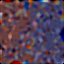}
  \includegraphics[width=0.30\linewidth]{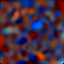}\\
\subcaption{The Interpolated images of Environment Colour}
 \end{center}
 \caption{Sample image of Environment colour captured by R1 - R3 rays and their interpolated images}
\label{fig:environmentColourSample}
\end{figure}

To estimate the environmental colour, we first hit the $H_{t}$ surface with rays from each pixel, consider techniques such as hemisphere sampling and we randomly reflect the rays toward the environment. As a result, the images of the environment are captured for the various levels/depths of ray reflection. In this study, we only use up to 3 reflection rays (1 to 3) with just a single sampling, as shown in figure \ref{fig:environmentColourExtraction}. Because we are not calculating all the ray reflections within the environment, we will have pixel locations without intensity values. An example can be seen in figure  \ref{fig:environmentColourSample}. Therefore, we are using a non-uniform interpolation algorithm \cite{Thvenaz1999ImageIA} to approximate the missing values in the obtained environmental intensity images $E^{r}_{t}$, where $r$ corresponds to the number of ray reflections.

In figure \ref{fig:environmentColourSample}, we see that the more ray reflects, the less bright the pixels become. The main reason behind this phenomenon is because of ray tracing algorithm and considering that the first ray $r1$ has more influence on the final pixel intensity than the ray $r3$. Therefore, when we have more ray reflections, the intensity of the pixels needs to be reduced, accordingly.

In stage 5, we generate the new input images $A_{t+1}=A_{t}-E^{r}_{t}$ by subtracting the environmental intensity reducing the inter-reflections from the original input images. There are three different sets of images for each ray reflection $r1$, $r2$ and $r3$.

\begin{figure}[ht]
\begin{center}
  \includegraphics[width=0.35\linewidth]{image/exampleForOriginalImage_Sphere1}
  \includegraphics[width=0.35\linewidth]{image/fixedSphereEvn_R1}\\
  \includegraphics[width=0.35\linewidth]{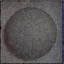}
  \end{center}
\caption{(Left) Image with inter-reflections, (Right) estimated environmental intensity image and (Bottom) obtained image without inter-reflections.}
\label{fig:differenceImagesSample}
\end{figure}

Finally, the obtained images which have fewer inter-reflections (example difference image is shown in figure \ref{fig:differenceImagesSample}) are used for as input to photometric stereo, generating a new $H_{t+1}$ surface. The whole process can be applied iteratively for a certain number of iterations or until the difference $D_{H}=H_{t+1}-H_{t}$ between a new 3D surface and the previous one is less than a given threshold.

\begin{figure}[ht]
\centering
  \includegraphics[width=0.3\linewidth]{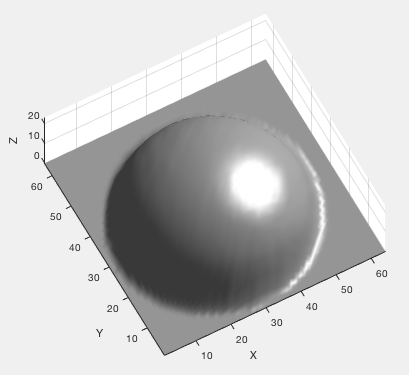}
  \includegraphics[width=0.3\linewidth]{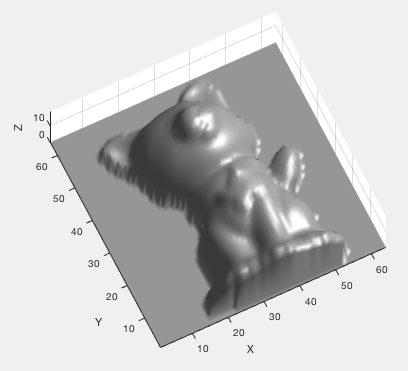}
  \includegraphics[width=0.3\linewidth]{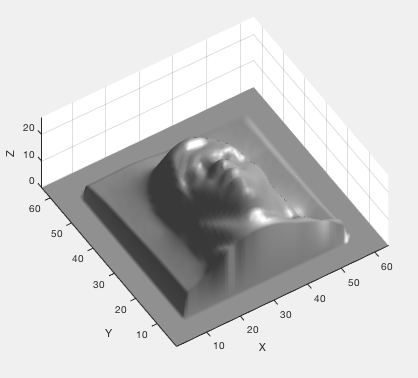}
\caption{Ground truth used for rendering and evaluation purpose. Synaptic Matlab Sphere, Harvard’s Photometric data, Scan Dat.}
\label{fig:groundTruthExample1}
\end{figure}

\begin{figure}[h!]
\centering
  \includegraphics[width=0.24\linewidth]{image/exampleForOriginalImage_Sphere1}
  \includegraphics[width=0.24\linewidth]{image/exampleForOriginalImage_Sphere3}
  \includegraphics[width=0.24\linewidth]{image/exampleForOriginalImage_Sphere4}
  \includegraphics[width=0.24\linewidth]{image/exampleForOriginalImage_Sphere1}\\
  \includegraphics[width=0.24\linewidth]{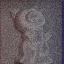}
  \includegraphics[width=0.24\linewidth]{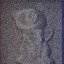}
  \includegraphics[width=0.24\linewidth]{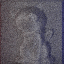}
  \includegraphics[width=0.24\linewidth]{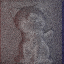}\\
  \includegraphics[width=0.24\linewidth]{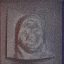}
  \includegraphics[width=0.24\linewidth]{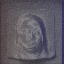}
  \includegraphics[width=0.24\linewidth]{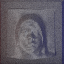}
  \includegraphics[width=0.24\linewidth]{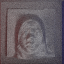}
\caption{Image samples with rendered inter-reflections.}
\label{chart:overallErrorChart}
\end{figure}

\section{Experiments and Results}
In our comparative evaluation study, three different datasets with ground truth were used. Scan data from the Harvard PS dataset \cite{3909}, a dataset with faces \cite{ArgPet2008} and synthetic data generated by simulated objects (see figures \ref{fig:groundTruthExample1} and \ref{chart:overallErrorChart}).

We used the photometric stereo approach to reconstruct the sets of the acquired $H_{t}$ surface, with and without inter-reflections considering different numbers (1 to 3) of ray reflections in the proposed reverted Monte-Carlo ray tracing algorithm. We then estimate the height-, albedo- and normal-error comparing to classic PS method \cite{Sun2007} using the available ground truth.

To calculate the height-error we used the equation,
\begin{equation}
 \overline{H}_{err} = \frac{1}{n}\Bigg(\sum_{i=1}^{n}|H_{GT} - H_{t}|_i\Bigg)
 \label{eq:heightError}
\end{equation}

$\overline{{H}}_{err}$ is the mean for height error. $H_{GT}$ is the height value of ground truth surface, whereas $H_t$ is the height value of reconstructed surface. Regarding the albedo-error we use the equation below,

\begin{equation}
\begin{split}
P_{err}^{r} = |P_{GT}^{r} - P_{H}^{r}| \\
P_{err}^{g} = |P_{GT}^{g} - P_{H}^{g}|\\
P_{err}^{b} = |P_{GT}^{b} - P_{H}^{b}| \\
P_{err}^{rgb} = \frac{\overline{P_{err}^{r}} + \overline{P_{err}^{g}}+\overline{P_{err}^{b}}}{3}
\end{split}
\label{eq:heightError}
\end{equation}

where $P_{err}^{rgb}$ is the albedo-error from mean of individual colour channel; Red $P_{err}^{r}$, Green $P_{err}^{g}$, and Blue $P_{err}^{b}$ channel.


Likewise, to calculate normal-error we utilise the following equation:
\begin{equation}
\begin{split}
N_{err}^{x} = |N_{GT}^{x} - N_{H}^{x}| \\
N_{err}^{y} = |N_{GT}^{y} - N_{H}^{y}|\\
N_{err}^{z} = |N_{GT}^{z} - N_{H}^{z}| \\
N_{err}^{xyz} = \frac{\overline{N_{err}^{x}} + \overline{N_{err}^{y}}+\overline{N_{err}^{z}}}{3}
\end{split}
\label{eq:heightError}
\end{equation}
$N_{err}^{xyz}$ denote the mean normal-error for all the axis $x,y,$ and $z$.Where $ \overline{N_{err}^{x}}$ is a mean error for X axis, $\overline{N_{err}^{y}}$ is mean error for Y, and $\overline{N_{err}^{z}}$ is mean error for Z, $N_{H}^{xyz}$ is normal from reconstructed surface.


\begin{figure}[ht]
\begin{center}
  \includegraphics[width=0.34\linewidth]{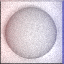}
 \includegraphics[width=0.34\linewidth]{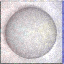}\\
 \includegraphics[width=0.34\linewidth]{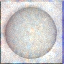}
 \includegraphics[width=0.34\linewidth]{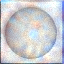}
 \end{center}
\caption{Example of the estimated albedo using classic PS \cite{Sun2007}, and the proposed IRT-PS method using 1-, 2- and 3-ray reflections.}
\label{fig:albedoErrorExample}
\end{figure}

\begin{table}[ht]
\begin{center}
\caption{Obtained results for the synthetic data, the Harvard and the face PS database comparing the \cite{Sun2007} method, with the 3 variations of the proposed IRT-PS approach.}
\begin{tabular}{ |l|c|c|c|c|c|} \hline
\textbf{Synthetic} & \textbf{PS} \cite{Sun2007} & \textbf{IRTPSr1} &  \textbf{IRTPSr2} & \textbf{IRTPSr3} \\\hline
\textbf{Height} &18.653 & 18.460 & 18.565 & \textbf{\textcolor{green}{18.436}}\\\hline
\textbf{Albedo} &0.082 & 0.082 & \textbf{\textcolor{green}{0.081}} & 0.087\\\hline
\textbf{Normal} &0.825 & 0.824 & 0.824 &  \textbf{\textcolor{green}{0.823}}\\\hline

\textbf{Harvard} & \textbf{PS} \cite{Sun2007} & \textbf{IRTPSr1}  & \textbf{IRTPSr2} &  \textbf{IRTPSr3} \\\hline
\textbf{Height}  &8.150 &8.140 &8.097 & \textbf{\textcolor{green}{7.296}}\\\hline
\textbf{Albedo} &0.522 & \textbf{\textcolor{green}{0.518}} &0.520&0.521\\\hline
\textbf{Normal}  &0.840 &0.839 & \textbf{\textcolor{green}{0.838}}	&0.840\\\hline

\textbf{Face} &  \textbf{PS} \cite{Sun2007} & \textbf{IRTPSr1}  &  \textbf{IRTPSr2} &  \textbf{IRTPSr3} \\\hline
\textbf{Height}  &9.341 &9.181 &9.272 & \textbf{\textcolor{green}{8.835}} \\\hline
\textbf{Albedo}  &0.235 &0.231 & \textbf{\textcolor{green}{0.230}} &0.241 \\\hline
\textbf{Normal}  &0.823  &0.823 &0.8221 & \textbf{\textcolor{green}{0.822}} \\ \hline \hline \hline

\textbf{Overall} & \textbf{PS} \cite{Sun2007} & \textbf{IRTPSr1}  &  \textbf{IRTPSr2} &  \textbf{IRTPSr3} \\\hline
\textbf{Height}  & \textbf{\textcolor{red}{12.049}} & 11.927 & 11.978&  \textbf{\textcolor{green}{11.523}}\\\hline
\textbf{Albedo}  & \textbf{\textcolor{red}{0.280}}	& \textbf{\textcolor{green}{0.2772}}	&0.2773	&0.283\\\hline
\textbf{Normal}  & \textbf{\textcolor{red}{0.829}} &0.829 & \textbf{\textcolor{green}{0.8283}}&	0.8288\\\hline
\end{tabular}
\end{center}
\label{tb:sphereError}
\end{table}

\begin{figure}[ht]
\begin{center}
  \includegraphics[width=0.475\linewidth]{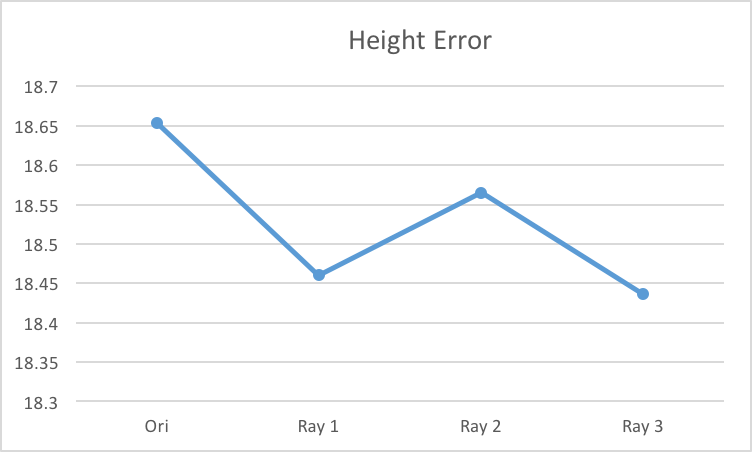}
  \includegraphics[width=0.475\linewidth]{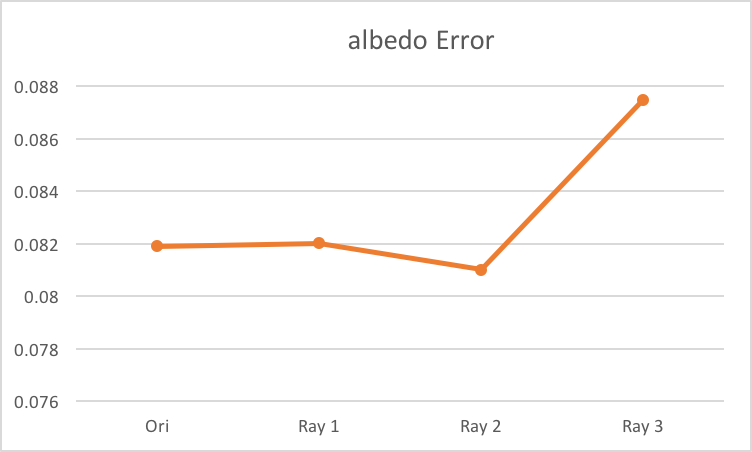}\\
  \includegraphics[width=0.475\linewidth]{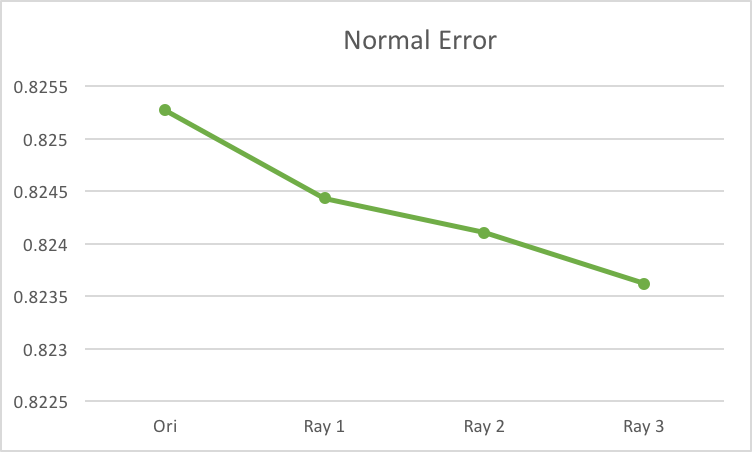}
   \subcaption{Overall Results for syntactic database: (left) Height Error, (Middle) Albedo Error (Right) Normal Error}
  \includegraphics[width=0.475\linewidth]{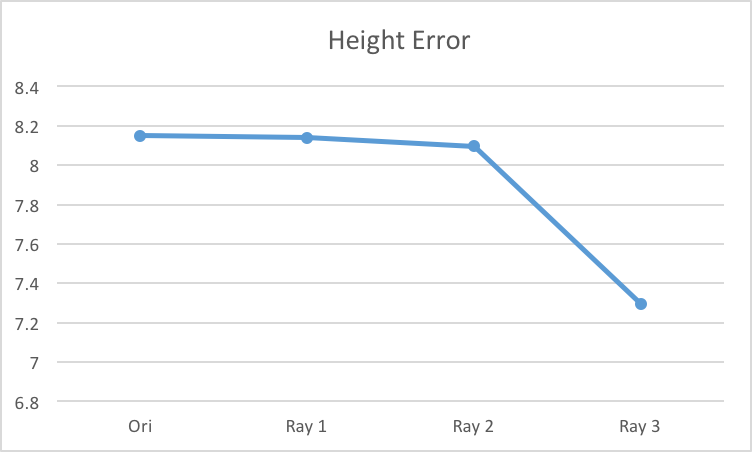}
  \includegraphics[width=0.475\linewidth]{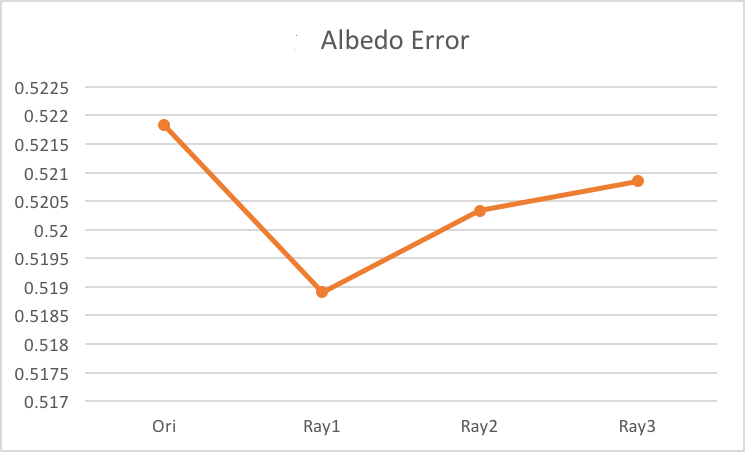}\\
  \includegraphics[width=0.475\linewidth]{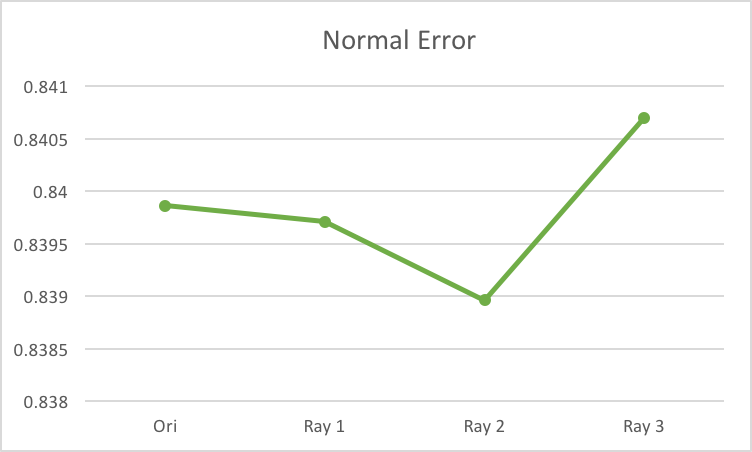}
\subcaption{Overall Results for face database \cite{ArgPet2008} :(left) Height Error, (Middle) Albedo Error (Right) Normal Error}
  \includegraphics[width=0.475\linewidth]{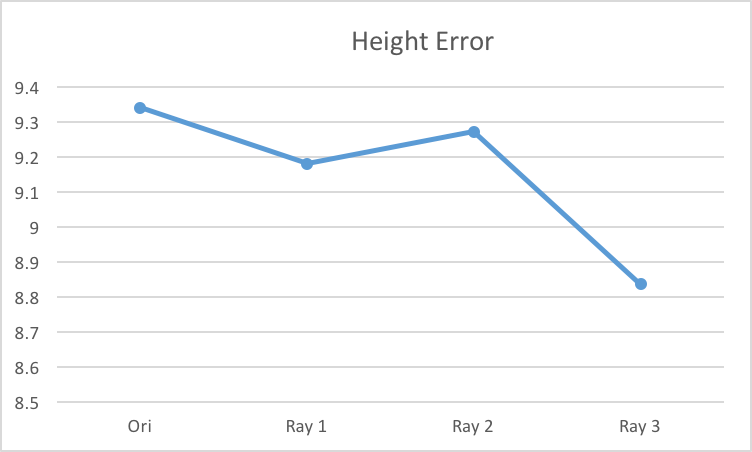}
  \includegraphics[width=0.475\linewidth]{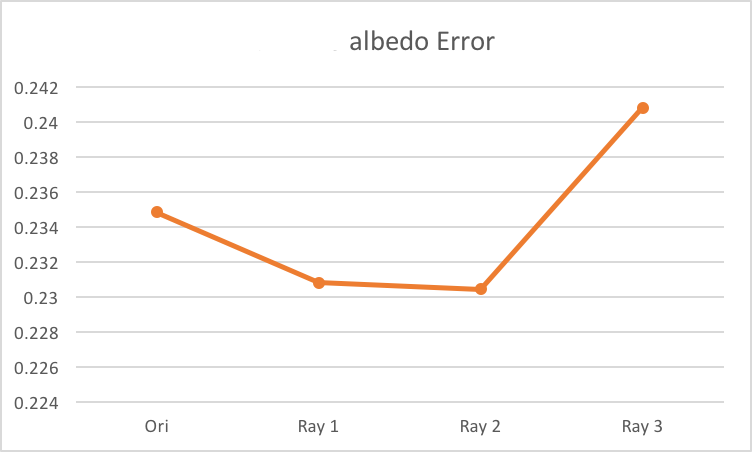}\\
  \includegraphics[width=0.475\linewidth]{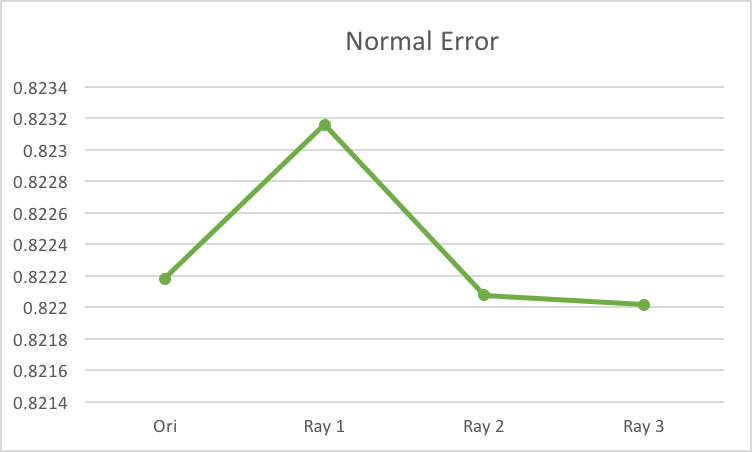}
  \subcaption{Overall Results for Harvard  database \cite{3909} : (left) Height Error, (Middle) Albedo Error (Right) Normal Error}
  \end{center}
  \caption{Overall results of the performed experiments demonstrating the r1 and r3 are the best methods for albedo and height estimation, respectively.}
\label{chart:overallErrorChart}
\end{figure}

From the table 1, and charts in figure \ref{chart:overallErrorChart}, we can see that the overall trend of mean Height, Albedo, and Normal errors are reduced with our approach than the classic photometric stereo one. In table 1, text highlighted in red are the average overall results of the \cite{Sun2007} photometric stereo method. Whereas best results from our IRT-PS approach are highlighted in the green text. From the charts figure \ref{chart:overallErrorChart}, we can see the general trend of the height error: Results improve with each additional ray and the best result is achieved by Ray 3. Likewise, the best result for Albedo and Normal are given by Ray 2. The indirect illumination captured by Rays R3 and R2 of the environment were able to reduce the inter-reflection effect from the original images. Furthermore,  looking at the overall table and comparing to PS \cite{Sun2007}, we again see that our method improves in all the estimation. The greatest improvement can be seen in Height, followed by Normal, and finally the Albedo error. This shows that if we improve the captured indirect illumination then it should result in more accurate and detailed reconstructed surfaces.

\section{Conclusions}
In this work, a novel iterative method considering inter-reflections both due to concavities and the environment was proposed. The IRT-PS approach iteratively applies Photometric Stereo and a reverted Monte-Carlo ray tracing algorithm, reconstructing the observed surface and separating the indirect from direct lighting. A comparative study was performed evaluating the reconstruction accuracy of the proposed solution on three different datasets and the overall results demonstrate improvement over the classic approaches that do not consider environmental inter-reflections.
\\\\
\section*{Acknowledgements}
\noindent This work is co-funded by the NATO within the WITNESS project under grant agreement number G5437. The Titan X Pascal used for this research was donated by the NVIDIA Corporation.

{\small
\bibliographystyle{ieee}
\bibliography{egbib}
}

\end{document}